\documentclass[sn-mathphys,Numbered]{sn-jnl}

\usepackage{graphicx}%
\usepackage{multirow}%
\usepackage{amsmath,amssymb,amsfonts}%
\usepackage{amsthm}%
\usepackage{mathrsfs}%
\usepackage[title]{appendix}%
\usepackage{xcolor}%
\usepackage{textcomp}%
\usepackage{adjustbox}%
\usepackage{manyfoot}%
\usepackage{booktabs}%
\usepackage{algorithm}%
\usepackage{algorithmicx}%
\usepackage{algpseudocode}%
\usepackage{listings}%
\floatname{algorithm}{Algorithm}     
\algrenewcommand\algorithmicrequire{\textbf{Input:}}
\algrenewcommand\algorithmicensure{\textbf{Output:}}

\theoremstyle{thmstyleone}%
\theoremstyle{thmstyletwo}%

\theoremstyle{thmstylethree}%

\begin{document}

\title[Adaptive recurrent flow-map operator learning for reaction–diffusion dynamics]{Adaptive recurrent flow-map operator learning for reaction–diffusion dynamics}


\author{\fnm{Huseyin} \sur{Tunc}}

\affil{\orgdiv{Department of Biostatistics and Medical Informatics}, \orgname{School of Medicine, Bahcesehir University},  \city{Istanbul}, \postcode{34744}, \country{Turkey}\\
huseyin.tunc@bau.edu.tr}


\abstract{Reaction-diffusion (RD) equations underpin pattern formation across chemistry, biology, and physics, yet learning stable operators that forecast their long-term dynamics from data remains challenging. Neural-operator surrogates provide resolution-robust prediction, but autoregressive rollouts can drift due to the accumulation of error, and out-of-distribution (OOD) initial conditions often degrade accuracy. Physics-based numerical residual objectives can regularize operator learning, although they introduce additional assumptions, sensitivity to discretization and loss design, and higher training cost. Here we develop a purely data-driven operator learner with adaptive recurrent training (DDOL-ART) using a robust recurrent strategy with lightweight validation milestones that early-exit unproductive rollout segments and redirect optimization. Trained only on a single in-distribution toroidal Gaussian family over short horizons, DDOL-ART learns one-step operators that remain stable under long rollouts and generalize zero-shot to strong morphology shifts across FitzHugh-Nagumo (FN), Gray-Scott (GS), and Lambda-Omega (LO) systems. Across these benchmarks, DDOL-ART delivers a strong accuracy and cost trade-off. It is several-fold faster than a physics-based numerical-loss operator learner (NLOL) under matched settings, and it remains competitive on both in-distribution stability and OOD robustness. Training-dynamics diagnostics show that adaptivity strengthens the correlation between validation error and OOD test error performance, acting as a feedback controller that limits optimization drift. Our results indicate that feedback-controlled recurrent training of DDOL-ART generates robust flow-map surrogates without PDE residuals, while simultaneously maintaining competitiveness with NLOL at significantly reduced training costs. }




\maketitle
\section{Introduction}\label{sec1}

Modeling the long-term evolution of spatiotemporal fields in reaction-diffusion (RD) systems is central to pattern formation, excitable media, and chemical kinetics, yet classical solvers become increasingly expensive as grids refine, stiffness grows, and forecast horizons extend \cite{Murray2003,hundsdorfer2003numerical}. Neural operator learning has emerged as a promising alternative by reducing simulation cost across families of partial differential equation (PDE) problems. Instead of computing a single trajectory, the goal is to learn an operator $\mathcal{G}$ that maps an input field to future solution states for many unseen inputs with a single forward pass \cite{Kovachki2023}. This paradigm stands in contrast to conventional coordinate-based neural solvers such as physics-informed neural networks (PINNs), which target one instance at a time via residual minimization \cite{Raissi2019,zhu2023bc}.

A first line of work represents the solution field directly as a continuous function of space-time (and optionally parameters), learning $u:\Omega\times[0,T]\to\mathbb{R}^m$ from scattered sensor measurements and query coordinates. Deep operator networks (DeepONets) implement a coordinate-conditioned operator model: a branch network encodes the input function via pointwise sensors, a trunk network encodes query coordinates $(\mathbf{x},t)$, and their inner product yields $u(\mathbf{x},t)$. Theoretical frameworks establish universal approximation of nonlinear continuous operators and discretization-invariant generalization across meshes \citep{Lu2021DeepONet, Kovachki2023}. Coordinate-based surrogates readily support arbitrary space-time query, parametric conditioning, and physics augmentation through residual penalties or weak-form losses \citep{Li2021_PINO}. Still, they can inherit the spectral bias of multi-layer perceptrons (MLPs) due to favoring low frequencies, which hampers the resolution of fine scales unless mitigated by architectural or encoding choices \citep{Rahaman2019_SpectralBias,Tancik2020_FourierFeatures}. In contrast to purely data-driven PINNs, operator-centric coordinate models (DeepONet) emphasize operator learning and discretization robustness. Fourier neural operator (FNO)-style kernels or higher-resolution physics constraints can be combined with coordinate trunks to improve zero-shot resolution transfer and stability in PDE families \citep{Li2021_PINO,Li2021_FNO}.

A second family aligns the learning task with numerical time integration by parameterizing a one-step solution operator (flow map) $\Phi_\theta:\mathbf{u}^n\mapsto\mathbf{u}^{n+1}$ on a chosen state representation, and then composing $\Phi_\theta$ autoregressively over long-horizons. This inductive bias encourages weight sharing across time steps, facilitates straightforward coupling with convolutional, spectral, or graph message-passing updates, and enables the natural integration of physics-based discrete residuals \citep{Brandstetter2022_MPNN,Li2021_FNO,Li2021_PINO}. Training may be purely data-driven via unrolling against trajectory data, or hybridized with physics-informed residual losses that enforce PDE consistency at finer resolutions than the supervision grid \citep{Li2021_PINO}. A central challenge is exposure bias. When flow-map models are trained with teacher forcing, each step conditions on ground-truth states, whereas at inference, the model must condition on its own predictions. Consequently, small per-step errors accumulate and can destabilize the PDE solution over long-horizons. Mitigating this error propagation requires recurrent or physics-constrained training strategies tailored to operator learning rather than generic sequence models \citep{geneva2020modeling}. Recent recurrent training strategies for neural operators explicitly align train/test dynamics and prove linear (rather than exponential) worst-case error growth, producing markedly improved long-term stability on canonical PDE benchmarks \citep{Ye2025_RNO}.

These two axes, representation (cartesian and flow map) and supervision (physics-informed and data-driven), yield complementary trade-offs. Physics-informed training reduces labeled-trajectory requirements and can regularize extrapolation by penalizing violations of the governing equations, yet it is sensitive to discretization choices, loss balancing, and how faithfully the residual encodes the true solution manifold \citep{Raissi2019,Li2021_PINO}. Data-driven training aligns seamlessly with the empirical solution operator and can effectively capture the unresolved effects inherent in high-fidelity data. However, out-of-distribution (OOD) generalization remains susceptible to fragility without the utilization of simulation-aligned curricula and inductive biases that preserve conservation laws or multiscale structure. Recent analyses have highlighted spectral bias and multiscale aliasing as fundamental failure modes for operator learners \citep{Liu2024_MsPDE_SpectralBias}. Large-scale geoscience case studies highlight both the promise and pitfalls of purely data-driven approaches. For example, adaptive FNO-based weather emulators achieve high-resolution skill and extreme speedups, yet long-horizon stability and error growth still depend on careful architectural and training design \citep{Pathak2022_FourCastNet,Kurth2022_FourCastNet_Scaling}.

Long-horizon stability and OOD robustness are contingent upon closing the train–test gap generated by teacher forcing. Recurrent unrolling, which enables the operator’s own predictions during training, aligns optimization with autoregressive inference. This approach provably transforms worst-case exponential error growth into linear, resulting in significantly more stable long-term rollouts \citep{Ye2025_RNO}. Fully discrete and scheme-consistent residual losses provide physics-aware objectives that can be utilized without trajectory labels and readily integrate with recurrent unrolling to enhance generalization \citep{Geng2024,geng2025end}. Notably, Geng \textit{et al.} developed ResNet operators trained on fully discrete residuals for Allen–Cahn dynamics and introduced an efficient sample-generation strategy throughout the time evolution \citep{Geng2024}. This robust training strategy is a pivotal component in our study.

While physics-based learning is often credited with improved extrapolation and robustness in PDE surrogates, our focus is on a complementary goal. We study whether a purely data-driven flow-map operator can achieve long-horizon stability and robustness to distribution shift when it is trained with an inference-aligned recurrent strategy that is explicitly regulated by feedback. We also study whether this can be achieved at substantially lower training cost. To this end, we examine three flow-map operator learners on periodic 2D reaction-diffusion benchmarks (FN, GS, LO) and address two primary challenges encountered in solving evolution PDEs: long-horizon stability and out-of-distribution (OOD) robustness. The first is a numerical-discretization-loss operator learner (NLOL) trained from a fully discrete numerical residual. The second is a supervised data-driven learner (DDOL) trained from numerical solution trajectories under free-run recurrent supervision. The third is DDOL-ART, an adaptive recurrent variant that builds on the robust recurrent training strategy of Geng \textit{et al.} \cite{Geng2024} and extends it with lightweight validation milestones that early-exit unproductive rollout segments, reset the rollout, and redirect optimization toward regimes that improve generalization. Across all systems, we use a stringent evaluation protocol and every trained model is rolled out autoregressively to a common test horizon. The reported gains reflect control of compounding error rather than short-horizon fit. Our results show that this adaptive controller yields a strong accuracy and cost trade-off. DDOL-ART is consistently faster than NLOL (\(3.2\times\) to \(3.6\times\) speedup) under matched settings and remains competitive in both in-distribution model selection and OOD generalization across the challenging morphology shifts considered in this work. More broadly, because DDOL-ART is developed as a data-driven recurrent training principle rather than a PDE-specific penalty, it points toward a transferable recipe for learning stable flow maps in general dynamical-systems forecasting from data, including settings where explicit governing equations are unavailable.

\section{Methodology}
In this section, we formalize the two-component reaction-diffusion dynamics on a periodic domain and outline the benchmark systems considered. We then describe the data pipeline, which combines high-order time integration with second-order spatial discretization, and curate initial conditions that span both ID and OOD domains. Next, we present a flow-map neural operator that learns a one-step update and is rolled out autoregressively, with an architecture that respects periodic boundaries. We then explain our NLOL, DDOL, and DDOL-ART approaches in detail. 

\subsection{Reaction–Diffusion Model Formulation}
A general two-component reaction–diffusion (RD) system on a periodic rectangular domain $\Omega=[0,1]^2$ can be written as
\begin{equation}\label{eq:RD-general}
\begin{aligned}
\partial_t u &= D_u \Delta u + R_u(u,v), \\
\partial_t v &= D_v \Delta v + R_v(u,v),
\end{aligned}
\end{equation}
where $(u,v)$ are the species concentrations, $D_u,D_v>0$ are diffusion coefficients, $\Delta$ is the Laplacian, and $(R_u,R_v)$ are nonlinear reaction terms \citep{rao2023encoding}. In all numerical experiments, we set $D_u=D_v=0.01$. We examine three prototypical systems.

Lambda-Omega (LO) RD system: The reaction kinetics are
\begin{equation}\label{eq:LO}
\begin{aligned}
R_u(u,v) &= (1 - u^2 - v^2)u + \beta (u^2 + v^2) v, \\
R_v(u,v) &= -\beta (u^2 + v^2) u + (1 - u^2 - v^2) v,
\end{aligned}
\end{equation}
with $\beta=1$ \citep{rao2023encoding}. LO is a normal-form model for oscillatory kinetics near a Hopf bifurcation. The radial term drives a unit-amplitude limit cycle while the $\beta$-coupled rotation sets the angular frequency, allowing target and spiral waves once diffusion is added \citep{Kopell1973}. It is widely used as a minimal oscillatory medium for studying wave selection and spiral dynamics in RD systems \citep{Kopell1973}.

FitzHugh-Nagumo (FN) RD system: The reaction kinetics are
\begin{equation}\label{eq:FN}
\begin{aligned}
R_u(u,v) &= u - u^3 - v + \alpha, \\
R_v(u,v) &= \beta (u - v),
\end{aligned}
\end{equation}
with $\alpha=0.01$ and $\beta=0.25$ \citep{rao2023encoding}. The FN model is a canonical excitatory medium model that accurately captures the rapid activation (cubic nonlinearity) and slow recovery processes of nerve impulses \citep{FitzHugh1961,Nagumo1962}. Originally proposed as a simplified version of the Hodgkin–Huxley model, the FN model has since been extensively studied. With the inclusion of diffusion, the FN model can simulate traveling pulses and wave trains. Rigorous analyses have identified families of traveling fronts, providing a comprehensive understanding of the intricate spatio-temporal behavior of the model \citep{deng1991existence,cebrian2024six}. 

Gray-Scott (GS) RD system: The reaction kinetics are
\begin{equation}\label{eq:GS}
\begin{aligned}
R_u(u,v) &= -u v^2 + F (1 - u), \\
R_v(u,v) &= u v^2 - (F + \kappa) v,
\end{aligned}
\end{equation}
with $F=0.025$ and $\kappa=0.055$ \citep{rao2023encoding}. The GS model describes cubic autocatalysis with feed/kill terms and generates diverse patterns such as spots, stripes, labyrinths and self-replicating spots \citep{gray1984autocatalytic,reynolds1997self}. 

\subsection{Data Generation}
For the time integration of the RD systems, we employ the third-order strong stability-preserving Runge-Kutta method (SSP-RK3). This method preserves the monotonicity and total variation diminishing (TVD) properties of the forward Euler step under an appropriate Courant–Friedrichs–Lewy (CFL) restriction, thereby delivering higher accuracy without introducing spurious oscillations \citep{gottlieb1998total}. This makes it well-suited for RD dynamics, where diffusion smooths and reactions can exhibit stiffness.

Applied component-wise to $(u,v)$, the SSP-RK3 update from $\mathbf{u}^n$ to $\mathbf{u}^{n+1}$ is
\begin{equation}\label{eq:ssp-rk3}
\begin{aligned}
\mathbf{k}_1 &= D \Delta \mathbf{u}^n + R(\mathbf{u}^n), &
\mathbf{u}^\ast &= \mathbf{u}^n + \Delta t \,\mathbf{k}_1,\\
\mathbf{k}_2 &= D \Delta \mathbf{u}^\ast + R(\mathbf{u}^\ast), &
\mathbf{u}^{\ast\ast} &= \tfrac{3}{4}\,\mathbf{u}^n + \tfrac{1}{4}\big(\mathbf{u}^\ast + \Delta t\,\mathbf{k}_2\big),\\
\mathbf{k}_3 &= D \Delta \mathbf{u}^{\ast\ast} + R(\mathbf{u}^{\ast\ast}), &
\mathbf{u}^{n+1} &= \tfrac{1}{3}\,\mathbf{u}^n + \tfrac{2}{3}\big(\mathbf{u}^{\ast\ast} + \Delta t\,\mathbf{k}_3\big),
\end{aligned}
\end{equation}
with fixed time step $\Delta t=10^{-4}$. We discretize the Laplacian using a second-order central finite-difference (FD2) stencil on a uniform periodic grid ($h=1/N$) as follows:
\[
\Delta u_{i,j} \;=\; \frac{u_{i+1,j}+u_{i-1,j}+u_{i,j+1}+u_{i,j-1}-4u_{i,j}}{h^2},
\]
where indices are taken modulo $N$ to enforce periodicity. This five-point stencil is $O(h^2)$ accurate and standard for parabolic PDEs. In our numerical experiments, we set \(N=128\). To create training data, each initial condition \(\textbf{u}^0\) derived from a specific distribution is advanced using the SSP-RK3+FD2 approach, resulting in data pairs \(\{\textbf{u}^n, \textbf{u}^{n+1}\}_{n=0}^{M-1}\).

\subsection{Initial Condition Distribution}
We employ periodic $128{\times}128$ grids and sample training initial conditions (ICs) from a single toroidal Gaussian distribution. We draw a center $\mathbf{c}\!\sim\!\mathcal{U}([0,1]^2)$ and a width $\sigma\!\in[0.05,0.20]$, and set the initial condition as the toroidal Gaussian profile
\[
g(\mathbf{x};\mathbf{c},\sigma)=\exp\!\Big(-\tfrac{d_{\mathbb{T}^2}(\mathbf{x},\mathbf{c})^2}{2\sigma^2}\Big),
\]
where $d_{\mathbb{T}^2}$ is the wrap–around distance on the unit torus defined in Appendix~A. For two–field systems $(u,v)$, we draw independent seeds and normalize ranges consistently. OOD initial conditions probe robustness via controlled departures from the training distribution, grouped as: compositions (multi-Gaussian and dot lattice), noise-perturbed seeds (noisy Gaussians and Turing noise), and geometric and dynamical triggers (patch, annulus, and stripes). All generators enforce periodic boundaries and fixed normalization. We provide all details of the OOD initial conditions in Appendix~A.

\subsection{ResNet Flow-Map Operator Learning Architecture}
The flow-map neural operator $\Phi_\theta: \mathbf{u}^n \mapsto \mathbf{u}^{n+1}$ is implemented as a residual convolutional neural network (ResNet) with periodic padding, inspired by its success in capturing hierarchical features in image-like data while maintaining stability through skip connections \cite{he2016deep}. Our study employs the ResNet structure, utilizing the hyperparameter configurations suggested by Geng et al. \cite{Geng2024}. The model processes inputs $\mathbf{u}^n \in \mathbb{R}^{b \times C \times H \times W}$ (mini-batch size $b$, channels $C=2$ for two-component RD systems, height/width $H=W=128$) as follows:
\[
\Phi_\theta(\mathbf{u}^n) = \mathcal{C}_\text{final} \circ \sigma \circ \mathcal{R}_4 \circ \sigma \circ \mathcal{R}_3 \circ \sigma \circ \mathcal{R}_2 \circ \sigma \circ \mathcal{R}_1 \circ \mathcal{C}_0(\mathbf{u}^n),
\]
where $\mathcal{C}_0$ is a $3\times3$ convolution lifting to mid-channels $C_{mid}=16$, each residual block $\mathcal{R}_k$ is
\[
\mathcal{R}_k(\mathbf{z}) = \mathbf{z} + \sigma(\mathcal{C}_{k+1} \circ \sigma(\mathcal{C}_k(\mathbf{z}))),
\]
with $3\times3$ convolutions $\mathcal{C}_{k}$ with stride one and circular padding to maintain size, \(tanh\) activation $\sigma$, and group normalization after each convolution layer for stability. The final $\mathcal{C}_\text{final}$ is a $3\times3$ convolution reducing back to $C$ channels without activation, allowing unrestricted output ranges. Periodic padding is applied via custom toroidal wrapping to enforce boundary conditions, preventing edge artifacts.

\subsection{Numerical Discretization Loss Operator Learning (NLOL)}
We introduce a physics-based operator learning approach called NLOL by utilizing space–time discretization for reaction–diffusion (RD) systems as proposed by Geng et al. \cite{Geng2024}. NLOL trains $\Phi_{\theta}$ by minimizing a Crank–Nicolson (CN) residual combined with second-order finite-difference (FD2) spatial discretization, without trajectory labels. For any given $\mathbf{u}^0 \in \mathbb{R}^{b \times H \times W}$ and $\mathbf{v}^0 \in \mathbb{R}^{b \times H \times W}$, the CN–FD2 scheme for the RD system \eqref{eq:RD-general} yields
\begin{equation}\label{eq:CN}
\begin{aligned}
\frac{\mathbf{u}^{n+1} - \mathbf{u}^n}{\Delta t} &= \frac{1}{2}\Big[D_u\,\Delta_h \mathbf{u}^{n} + R_u(\mathbf{u}^{n},\mathbf{v}^{n})
                 + D_u\,\Delta_h \mathbf{u}^{n+1} + R_u(\mathbf{u}^{n+1},\mathbf{v}^{n+1})\Big],\\
\frac{\mathbf{v}^{n+1} - \mathbf{v}^n}{\Delta t} &= \frac{1}{2}\Big[D_v\,\Delta_h \mathbf{v}^{n} + R_v(\mathbf{u}^{n},\mathbf{v}^{n})
                 + D_v\,\Delta_h \mathbf{v}^{n+1} + R_v(\mathbf{u}^{n+1},\mathbf{v}^{n+1})\Big],
\end{aligned}
\end{equation}
with time step $\Delta t$ and the standard five-point FD2 Laplacian
$\Delta_h u_{i,j}=\big(u_{i+1,j}+u_{i-1,j}+u_{i,j+1}+u_{i,j-1}-4u_{i,j}\big)/h^2$ on a uniform grid of spacing $h$.
We draw an initial condition $\mathbf{U}^0\in\mathbb{R}^{b\times 2\times H\times W}$ from the training distribution and use only initial-condition data, i.e., we set $\hat{\mathbf{U}}^{\,0}=\mathbf{U}^{0}$. Neural-operator predictions are denoted by $\hat{\mathbf{U}}^{\,n}\in\mathbb{R}^{b\times 2\times H\times W}$ with stacked components $\hat{\mathbf{U}}^{\,n}=\big(\hat{\mathbf{u}}^{\,n},\hat{\mathbf{v}}^{\,n}\big)^{\top}$ with
$\hat{\mathbf{u}}^{\,n},\hat{\mathbf{v}}^{\,n}\in\mathbb{R}^{b\times H\times W}$, and they satisfy the one-step flow map
$\hat{\mathbf{U}}^{\,n+1}=\Phi_{\theta}\!\big(\hat{\mathbf{U}}^{\,n}\big)$.
Let $\mathbf{D}=\mathrm{diag}(D_u,D_v)$ act channel-wise, and define the reaction operator
$\mathbf{R}(\hat{\mathbf{U}})=\big(R_u(\hat{\mathbf{u}},\hat{\mathbf{v}}),\,R_v(\hat{\mathbf{u}},\hat{\mathbf{v}})\big)^{\top}$.
We compactly write the CN residual as
\begin{equation}\label{eq:CN_residual}
\mathcal{F}\!\big(\hat{\mathbf{U}}^{\,n+1},\hat{\mathbf{U}}^{\,n}\big)
= \hat{\mathbf{U}}^{\,n+1}-\hat{\mathbf{U}}^{\,n}
-\frac{\Delta t}{2}\Big[\mathbf{D}\,\Delta_h\hat{\mathbf{U}}^{\,n}+\mathbf{R}\!\big(\hat{\mathbf{U}}^{\,n}\big)
+\mathbf{D}\,\Delta_h\hat{\mathbf{U}}^{\,n+1}+\mathbf{R}\!\big(\hat{\mathbf{U}}^{\,n+1}\big)\Big],
\end{equation}
and train $\Phi_{\theta}$ by minimizing the discrete numerical loss

\begin{equation}\label{eq:NLOLLoss}
\mathcal{L}_{\mathrm{NLOL}}^{\,n}
=\frac{1}{4bHW}\,
\big\|\mathcal{F}\!\big(\hat{\mathbf{U}}^{\,n+1},\hat{\mathbf{U}}^{\,n}\big)\big\|_{F}^{2},
\end{equation}

where $\|\cdot\|_{F}$ denotes the Frobenius norm over the stacked $(b,2,H,W)$ dimensions, and $\Delta_h$ is applied component-wise to $\hat{\mathbf{u}}$ and $\hat{\mathbf{v}}$. This NLOL approach is physics-based and does not require data except that the initial condition $\mathbf{U}^0$. The critical issue is how to use the local discrete loss $\mathcal{L}_{\mathrm{NLOL}}^{\,n}$ with a robust learning strategy to obtain a reliable flow-map operator $\Phi_{\theta}$. We explain the robust learning strategy in Section \ref{sec2_7}.

\subsection{Data-Driven Training with Recurrent Supervision (DDOL)}
DDOL is a data-driven counterpart to NLOL that trains the flow-map operator $\Phi_{\theta}$ from numerically generated trajectories, aligning training with inference by always rolling out on the model’s own predictions (no teacher forcing). Let $\Psi_{\mathrm{SSP\text{-}RK3,FD2}}$ denote a reference solver advancing one time step with strong-stability-preserving third-order Runge–Kutta in time and a five-point second-order finite difference Laplacian in space. Given a total batch of $B$ trajectories
\[
\mathcal{D}=\big\{\,(\mathbf{U}_k^0,\mathbf{U}_k^1,\ldots,\mathbf{U}_k^T)\ \big|\ \mathbf{U}_k^{n+1}=\Psi_{\mathrm{SSP\text{-}RK3,FD2}}\!\left(\mathbf{U}_k^{n};\Delta t,h\right),\ k=1,\ldots,B \big\},
\]
each state is channel-stacked as $\mathbf{U}_k^n=(\mathbf{u}_k^n,\mathbf{v}_k^n)^{\top}\in\mathbb{R}^{2\times H\times W}$. At training time, we form mini-batches of size $b$ by stacking $b$ trajectories along a batch axis, so that $\mathbf{U}^{n}\in\mathbb{R}^{b\times 2\times H\times W}$. Thus, we have \(B=pb\) where \(p\) represents the number of mini-batches. For any mini-batch, we initialize the rollout with the ground-truth initial state and then unroll purely on predictions,
\[
\hat{\mathbf{U}}^{\,0}=\mathbf{U}^{0},\qquad \hat{\mathbf{U}}^{\,n+1}=\Phi_{\theta}\!\big(\hat{\mathbf{U}}^{\,n}\big),\quad n=0,1,\ldots,
\]
while supervising each predicted next state against the reference solver’s target $\mathbf{U}^{\,n+1}$. Using a mean-squared error over channels and spatial dimensions, the step loss reads

\begin{equation}\label{eq:DDOL_step}
\mathcal{L}_{\mathrm{DDOL}}^{\,n}
=\frac{1}{4bHW}\,
\big\|\,\Phi_{\theta}\!\big(\hat{\mathbf{U}}^{\,n}\big)-\mathbf{U}^{\,n+1}\,\big\|_{F}^{2},
\end{equation}
where $\|\cdot\|_{F}$ is the Frobenius norm over the stacked $(b,2,H,W)$ dimensions. This recurrent supervision avoids the train–test mismatch of teacher forcing by conditioning on the model’s own predictions at every step, while the supervision targets come from the high-fidelity SSP–RK3+FD2 integrator.

\subsection{Recurrent Training Strategy and Adaptivity}\label{sec2_7}
We train the one–step flow map $\Phi_{\theta}$ by recurrently rolling out on the model's own predictions over a window and updating parameters at every step. This aligns training with inference and avoids the train–test mismatch of teacher forcing (exposure bias) documented in sequence learning and recent neural operator studies. Concretely, the model always conditions on $\hat{\mathbf{U}}^{\,n}$ to predict $\hat{\mathbf{U}}^{\,n+1}=\Phi_{\theta}(\hat{\mathbf{U}}^{\,n})$, and gradients are taken with respect to the step loss. We adopt the robust learning strategy provided by Geng et al. \cite{Geng2024} for our NLOL and DDOL models, and the pseudocode is given in Algorithm~\ref{alg:recurrent}. This free-run supervision is analogous to recurrent operator training, which has been shown to improve long-horizon stability compared to teacher forcing \cite{Ye2025_RNO}. 

We assemble $B=pb$ initial conditions $\{\mathbf{U}_0^{(l)}\}_{l=1}^{B}$ and split them into $p$ disjoint mini-batches of size $b$. Stacking along the batch axis yields $\mathbf{U}^{0}\in\mathbb{R}^{b\times 2\times H\times W}$. For each subset, we start from the ground-truth initial state and rollout for a horizon of $M$ steps with fixed increment $\Delta t$. Thus, our final training time is \(T=M\Delta t\). At step $n$, we take $b_n$ inner optimizer updates on the current loss, then feed the new prediction forward by setting $\hat{\mathbf{U}}^{\,n}\leftarrow \hat{\mathbf{U}}^{\,n+1}$. This is the key step for the current training strategy and it differs from the recurrent neural operator training provided by Ye et. al. \cite{Ye2025_RNO}. The flow-map model first learns to better predict $\hat{\mathbf{U}}^{\,n+1}$ using $\hat{\mathbf{U}}^{\,n}$ and then used as input to compute $\hat{\mathbf{U}}^{\,n+2}=\Phi_{\theta}(\hat{\mathbf{U}}^{\,n+1})$.

For NLOL, the step loss is the CN–FD2 residual from \eqref{eq:NLOLLoss} evaluated on $(\hat{\mathbf{U}}^{\,n},\hat{\mathbf{U}}^{\,n+1})$ and needs no trajectory labels. For DDOL, supervision uses solver targets from the SSP–RK3+FD2 integrator, with the per-step MSE in \eqref{eq:DDOL_step}. In both cases the norm is Frobenius over $(b,2,H,W)$ and $\Delta_h$ acts component-wise. Algorithm~\ref{alg:recurrent} provides the robust training strategy step by step for NLOL and DDOL approaches.

While DDOL attains strong long-horizon accuracy by free-run supervision, its fixed rollout horizon $M$ can be computationally expensive and, when $M$ is large, prone to error accumulation. To reduce training cost while controlling long-time error growth, we adopt an adaptive variant, DDOL-ART, which interleaves the recurrent rollout with periodic, supervised validation on a small held-out set. At regular milestones (every $r=\lfloor M/10 \rfloor$ steps by default), the model is evaluated on validation pairs $(\mathbf{U}_{\mathrm{val}}^{\,n},\mathbf{U}_{\mathrm{val}}^{\,n+1})$ produced by the same SSP–RK3+FD2 generator as the training targets. If the current validation loss does not improve over the best so far, we increase a failure counter. Upon  \(n_{fail}\) consecutive validation non-improvements, we early-exit the current rollout and move to a fresh mini-batch of initial conditions. In practice, this adaptivity curbs long-time drift, reduces wasted compute on unproductive segments, and shortens wall-clock training while preserving the free-run alignment of training with inference.

Formally, let $\mathcal{V}=\{(\mathbf{U}_{\mathrm{val}}^{\,n},\mathbf{U}_{\mathrm{val}}^{\,n+1})\}_{n=0}^{M-1}$ be a small supervised validation set of $K\in\{4,8\}$ initial conditions with one-step targets from the SSP–RK3+FD2 solver. We consider \(B=8K\), i.e. \((B,K)=(32,4)\) or \((B,K)=(64,8)\). DDOL-ART follows the DDOL loop but augments it with a validation milestone interval $r=\lfloor M/10\rfloor$, a running best validation loss $J_{\text{best}}$, a failure counter $n_{fail}$, and model checkpointing when validation improves. At each step $n$, after $b_n$ inner updates on $\mathcal{L}_{\mathrm{DDOL}}^{\,n+1}$, we compute the validation loss whenever $(n\!+\!1)\bmod r=0$ as follows

\begin{equation}\label{eq:Jval}
\mathcal{J}_{\mathrm{val}}
=\max_{0\le m \le M-1}
\frac{1}{4KHW}\,
\big\|\Phi_{\theta}\!\big(\mathbf{U}_{\mathrm{val},k}^{\,m}\big)-\mathbf{U}_{\mathrm{val},k}^{\,m+1}\big\|_{F}^{2}.
\end{equation}

If $\mathcal{J}_{\mathrm{val}}<J_{\text{best}}$, we save the model and set $J_{\text{best}}\!\leftarrow\!\mathcal{J}_{\mathrm{val}}$ and failure count to zero ($c_{fail}\!\leftarrow\!0$). If $\mathcal{J}_{\mathrm{val}} \geq J_{\text{best}}$, we set $c_{fail}\!\leftarrow\!c_{fail}+1$. When $c_{fail}$ reaches to failure threshold $n_{fail}$, we stop the current rollout early and proceed to the next subset. The full procedure is summarized in Algorithm~\ref{alg:ddolart}.

We use a decaying inner-update budget $b_n$, and a learning-rate schedule that decays after each subset to fine-tune across different initial conditions, following the spirit of fully-discrete operator training \cite{Geng2024}. Concretely in our experiments, we use Adam with initial step size $\eta_0=10^{-3}$ and a per-subset decay, mini-batch size $b=4$, sample size of initial conditions $B \in \{32,64\}$, final time $T \in \{1,5\}$, model time increment $\Delta t \in \{0.005,0.01,0.02,0.05,0.1\}$ and $b_n$ linearly decreasing from $500$ to $100$. The entire training procedure explained for NLOL, DDOL and DDOL-ART in Algorithms~\ref{alg:recurrent} and \ref{alg:ddolart} is repeated twice within an outer loop. Two outer epochs were observed to be sufficient for stabilizing the validation and test errors as later discussed through Figure~\ref{fig:gs_training_dynamics} in numerical experiments.

\begin{algorithm}
\small
\caption{Training strategy of NLOL and DDOL}\label{alg:recurrent}
\begin{algorithmic}
\Require $\{\mathcal{B}_\ell\}_{\ell=1}^{p}$ (mini-batches of $b=4$ ICs and \(B=4p\)), horizon $M$, inner budgets $\{b_n\}_{n=0}^{M-1}$, optimizer Adam, model $\Phi_{\theta}$
\For{$epoch=1$ to $2$} \Comment{outer epoch}
\For{$\ell=1$ to $p$} \Comment{loop over mini-batches}
  \State Initialize rollout with ground-truth ICs: $\hat{\mathbf{U}}^{\,0} \gets \mathcal{B}_\ell=\mathbf{U}^{0}\!\in\!\mathbb{R}^{b\times 2\times H\times W}$
  \For{$n=0$ to $M-1$}

  \For{$t=1$ to $b_n$} \Comment{inner optimization at the same step}
    \State Predict next state: $\hat{\mathbf{U}}^{\,n+1}\gets \Phi_{\theta}(\hat{\mathbf{U}}^{\,n})$
    \If{NLOL} \Comment{physics-based}
       \State $\mathcal{L}\ \gets\ \mathcal{L}_{\mathrm{NLOL}}^{\,n}=\frac{1}{4bHW}\,\big\|\mathcal{F}\!\big(\hat{\mathbf{U}}^{\,n+1},\hat{\mathbf{U}}^{\,n}\big)\big\|_{F}^{2}$ \ \ (use \eqref{eq:NLOLLoss})
    \ElsIf{DDOL} \Comment{data-driven}
       \State $\mathcal{L}\ \gets\ \mathcal{L}_{\mathrm{DDOL}}^{\,n}=\frac{1}{4bHW}\,\big\|\,\Phi_{\theta}\!\big(\hat{\mathbf{U}}^{\,n}\big)-\mathbf{U}^{\,n+1}\,\big\|_{F}^{2}$ (use \eqref{eq:DDOL_step})
    \EndIf
    
       \State $\theta \gets \mathsf{optimizer}.\texttt{step}\big(\nabla_{\theta}\mathcal{L}\big)$ \ \ \ (recompute $\mathcal{L}$ after each step)
    \EndFor
    \State Feed forward: $\hat{\mathbf{U}}^{\,n}\gets \hat{\mathbf{U}}^{\,n+1}$
  \EndFor
  \State Decay the learning rate and proceed to the next subset
\EndFor
\EndFor
\end{algorithmic}
\end{algorithm}

\begin{algorithm}
\small
\caption{Training strategy of DDOL-ART}\label{alg:ddolart}
\begin{algorithmic}
\Require $\{\mathcal{B}_\ell\}_{\ell=1}^{p}$ (mini-batches of $b$ ICs and $B=4p$), horizon $M$, budgets $\{b_n\}_{n=0}^{M-1}$, validation set $\mathcal{V}$ of $K$ one-step pairs, optimizer Adam, model $\Phi_{\theta}$
\State Set milestone interval $r=\lfloor M/10\rfloor$ and failure threshold \(n_{fail}=2\); initialize $J_{\text{best}}\gets+\infty$, $c_{fail}\gets 0$
\For{$epoch=1$ to $2$} \Comment{outer epoch}
\For{$\ell=1$ to $p$} \Comment{loop over mini-batches}
  \State Initialize rollout: $\hat{\mathbf{U}}^{\,0} \gets \mathcal{B}_\ell=\mathbf{U}^{0}\in\mathbb{R}^{b\times 2\times H\times W}$
  \For{$n=0$ to $M-1$} \Comment{free-run rollout on predictions}
    \State Predict: $\hat{\mathbf{U}}^{\,n+1}\gets \Phi_{\theta}(\hat{\mathbf{U}}^{\,n})$
    \For{$t=1$ to $b_n$} \Comment{inner optimization at step $n$}
       \State $\mathcal{L}\gets \mathcal{L}_{\mathrm{DDOL}}^{\,n}=\frac{1}{4bHW}\,\big\|\,\Phi_{\theta}\!\big(\hat{\mathbf{U}}^{\,n}\big)-\mathbf{U}^{\,n+1}\,\big\|_{F}^{2}$
       \State $\theta \gets \mathsf{optimizer}.\texttt{step}\big(\nabla_{\theta}\mathcal{L}\big)$ \Comment{Adam update }
    \EndFor
    \State Feed forward: $\hat{\mathbf{U}}^{\,n}\gets \hat{\mathbf{U}}^{\,n+1}$
    \If{$(n+1)\bmod r=0$} \Comment{validation milestone}
       \State Compute $\displaystyle \mathcal{J}_{\mathrm{val}}$ using \eqref{eq:Jval}
       \If{$\mathcal{J}_{\mathrm{val}}<J_{\text{best}}$} \State Save checkpoint; $J_{\text{best}}\gets \mathcal{J}_{\mathrm{val}}$; $c_{fail}\gets 0$
       \Else \State $c_{fail}\gets c_{fail}+1$
       \EndIf
       \If{$c_{fail} = n_{fail}$} \State \textbf{break} \Comment{early-exit current rollout and go to next mini-batch}
       \EndIf
    \EndIf
  \EndFor
  \State Decay learning rate and proceed to the next subset
\EndFor
\EndFor
\end{algorithmic}
\end{algorithm}

\section{Numerical Experiments}\label{sec:experiments}

We begin by selecting models through an ID hyperparameter search, and then fix these choices for all subsequent evaluations. Specifically, for each benchmark (FN, GS, and LO), we train NLOL, DDOL, and DDOL-ART in horizons \(T\in\{1,5\}\), batch sizes \(B\in\{32,64\}\), and time increments \(\Delta t\in\{0.005,0.01,0.02,0.05,0.1\}\). After meticulously benchmarking various scenarios for ID test performance, we freeze these selections and retain them unchanged for the OOD tests and the long-horizon error growth study. We compare the training cost of operator learners and experimentally demonstrate how DDOL-ART produces correlated behavior between ID validation error and OOD test error. In the concluding section, we present qualitative contour reconstructions, underscoring how DDOL-ART tracks OOD dynamics across distinct initial condition families. All experiments and timing measurements reported in this work were conducted on a single workstation equipped with an AMD Ryzen Threadripper 2990WX 32-core CPU (3.0\,GHz), an NVIDIA GeForce RTX 3090 GPU with 24\,GB memory, and 32\,GB RAM.

\begin{figure}[t]
  \centering
  \includegraphics[width=\linewidth]{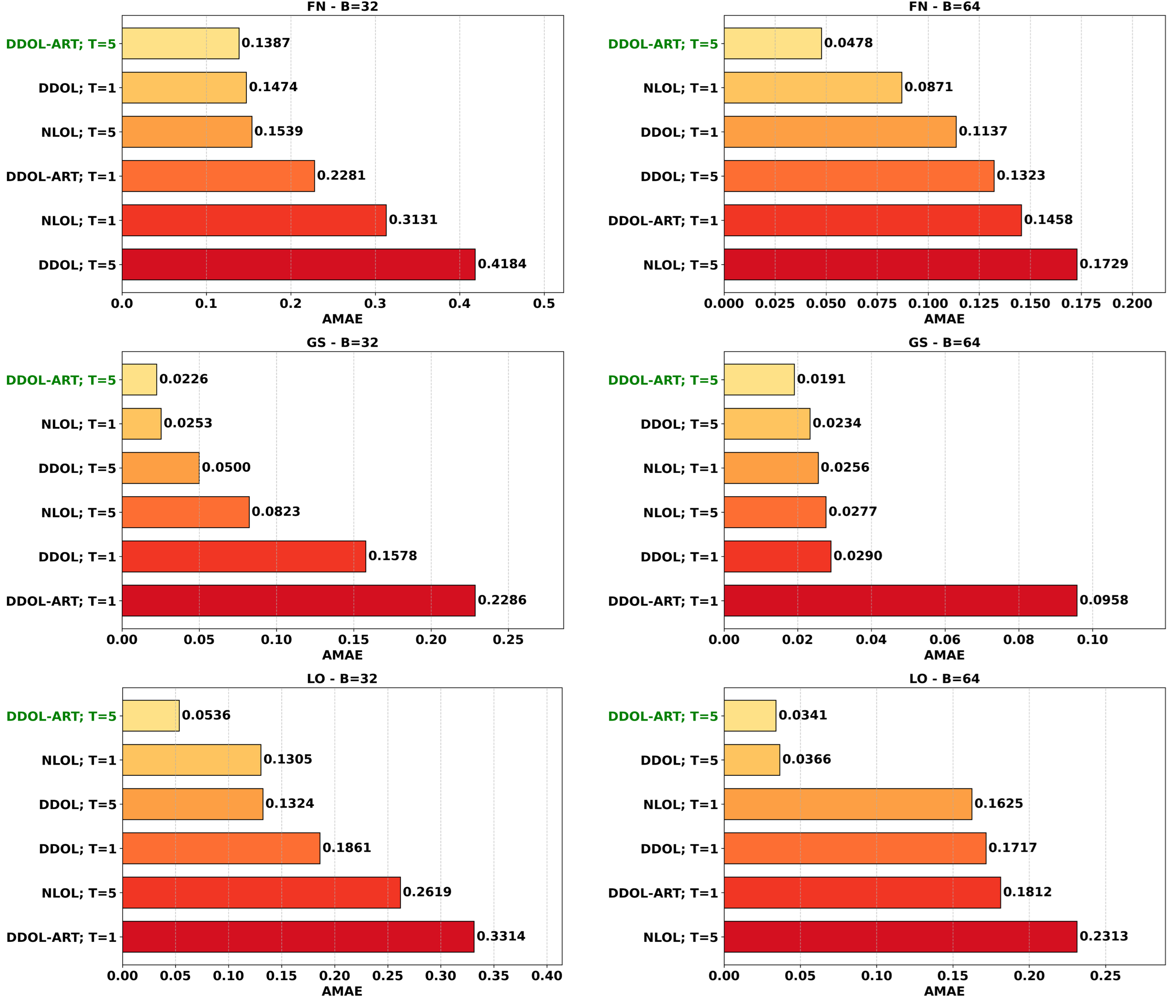}
  \caption{ID performance of the operator learning models. For each benchmark FN, GS, and LO, bars show the \(\Delta t\)-averaged AMAE obtained by NLOL, DDOL, and DDOL-ART models with \(B\in\{32,64\}\), \(T\in\{1,5\}\) and \(\Delta t\in\{0.005,0.01,0.02,0.05,0.1\}\). We considered \(10\) random seeds to generate ID initial conditions and measured AMAE using the final test time \(T=10\) for each model. The presented values are the average of the AMAE values of five different models trained using \(\Delta t\in\{0.005,0.01,0.02,0.05,0.1\}\). Entries are sorted in ascending order, with the best in each subplot highlighted in green.}
  \label{fig:hyper_tuning_amae}
\end{figure}

For each equation (FN, GS, LO), method (NLOL, DDOL, DDOL-ART), batch size \(B\in\{32,64\}\), and training horizon \(T\in\{1,5\}\), we train five models with \(\Delta t\in\{0.005,0.01,0.02,0.05,0.1\}\). For each configuration, we sample the initial conditions \(10\) times, autoregressively roll the model to a fixed evaluation horizon \(T_{\text{test}}=10\) using its native \(\Delta t\), compute the maximum absolute error (MAE) throughout the rollout against the SSP-RK3+FD2 reference for each seed, and then average these ten maxima to obtain the average MAE (AMAE) corresponding to that \(\Delta t\). In Fig.~\ref{fig:hyper_tuning_amae}, the bar for a given scenario represents the arithmetic mean of these five AMAE values across the five \(\Delta t\) replicas.

Figure~\ref{fig:hyper_tuning_amae} reports the in-distribution (ID) model-selection results for FN, GS, and LO under two batch sizes: \(B=32\) and \(B=64\). Here \(T\in\{1,5\}\) denotes the training unrolling horizon (the final time used during recurrent optimization), whereas every trained model is evaluated by autoregressively rolling out to a common test horizon \(T_{\text{test}}=10\) to compute AMAE. Thus, differences across \(T\) reflect how the training protocol shapes stability and compounding-error control under long rollouts, not an easier or shorter evaluation.

Across all six panels, DDOL-ART trained with \(T=5\) achieves the lowest \(\Delta t\)-averaged AMAE, indicating that adaptive recurrent supervision provides the most reliable ID configuration when performance is judged at \(T_{\text{test}}=10\). At \(B=32\), the gains over the next-best configuration are consistent but modest (FN: \(0.1387\) vs \(0.1530\); GS: \(0.0226\) vs \(0.0240\); LO: \(0.0536\) vs \(0.0549\)). At \(B=64\), the improvements become substantially larger (FN: \(0.0478\) vs \(0.0933\); GS: \(0.0191\) vs \(0.0300\); LO: \(0.0341\) vs \(0.1717\)), suggesting that larger batches reduce gradient noise in free-run training and make the benefit of the adaptive milestones more pronounced.

A mechanistic explanation for why DDOL-ART benefits from \(T=5\) (relative to \(T=1\)) follows directly from the evaluation protocol. 
Because AMAE is measured after a \(T_{\text{test}}=10\) rollout, the dominant failure mode is compounding error under repeated composition of the learned flow map. Training with \(T=5\) exposes the model to longer free-run trajectories during optimization, so it must learn to operate on its own slightly off-manifold predictions and implicitly correct drift in amplitude, phase, and morphology induced by the nonlinear reaction terms and diffusion. In contrast, \(T=1\) predominantly emphasizes local one-step accuracy near ground-truth states, which may not sufficiently penalize multi-step instability that only becomes apparent under long rollouts. The ART milestones further stabilize this longer-horizon training by early-exiting unproductive segments, preventing error amplification during optimization and yielding a better-calibrated operator for \(T_{\text{test}}=10\).

Two additional patterns are noteworthy. First, longer unrolling without adaptivity can be counterproductive for purely data-driven training: DDOL at \(T=5\) collapses for GS at \(B=64\) (AMAE \(0.4299\)), illustrating that multi-step drift during training can degrade the learned one-step map even when supervised targets are available. Second, the physics-based NLOL baseline can be competitive in some short-horizon regimes (e.g., GS at \(T=1\)), yet it often deteriorates when the training unrolling horizon increases, indicating that minimizing a discrete residual alone does not necessarily calibrate the composed dynamics required for stable rollouts to \(T_{\text{test}}=10\).

Based on the ID results in Fig.~\ref{fig:hyper_tuning_amae}, we fix the best-performing configuration for each PDE and model, and we keep these choices unchanged for all subsequent OOD and long-horizon experiments (Table~\ref{tab:model_selection}). This design prevents any post-hoc tuning on shifted test distributions and ensures that performance differences reported later reflect genuine generalization rather than re-optimization. Importantly, the selected configurations consistently favor the longer training unrolling horizon \(T=5\), which supports the interpretation that training on longer free-run compositions improves rollout stability when all methods are evaluated at the common horizon \(T_{\text{test}}=10\).  Finally, the selected operator step sizes \(\Delta t\) depend on both the PDE and the training strategy; for FN and GS the best DDOL and DDOL-ART variants tend to prefer larger \(\Delta t\), whereas LO prefers smaller steps, most notably for DDOL-ART, indicating that the optimal time increment is system-dependent and closely tied to recurrent-training stability.

\begin{table}[t]
\centering
\caption{Selected hyperparameters (\(\Delta t\), \(B\), \(T\)) for each PDE and operator learner, chosen to minimize the in-distribution AMAE.}\label{tab:model_selection}
\begin{tabular}{llccc}
\toprule
\textbf{PDE} & \textbf{Model} & \(\boldsymbol{\Delta t}\) & \(\boldsymbol{B}\) & \(\boldsymbol{T}\) \\
\midrule
FN & NLOL     & 0.02  & 64 & 5 \\
GS & NLOL     & 0.02  & 64 & 5 \\
LO & NLOL     & 0.02  & 64 & 5 \\
\midrule
FN & DDOL     & 0.10  & 64 & 5 \\
GS & DDOL     & 0.10  & 64 & 5 \\
LO & DDOL     & 0.02  & 64 & 5 \\
\midrule
FN & DDOL-ART & 0.10  & 64 & 5 \\
GS & DDOL-ART & 0.05  & 32 & 5 \\
LO & DDOL-ART & 0.005 & 64 & 5 \\
\bottomrule
\end{tabular}
\end{table}

Table~\ref{tab:training_time_hours} highlights the computational advantage of DDOL-ART compared to NLOL and DDOL operator learners. To ensure a fair comparison, all training times are measured under identical settings for every method, namely \(T=1\), \(B=32\), and \(\Delta t=0.01\), so the reported differences reflect the training strategy rather than a more favorable hyperparameter choice. Across all three RD systems, DDOL-ART consistently reduces wall-clock time from roughly \(2.2\) to \(2.4\) hours for NLOL to about \(0.63\) to \(0.71\) hours, corresponding to an approximately \(3.2\times\) to \(3.6\times\) speedup. Compared to standard DDOL, DDOL-ART remains faster by about \(1.8\times\) to \(2.2\times\), indicating that the adaptive recurrent mechanism eliminates unproductive optimization segments while preserving the free-run alignment between training and inference.

\begin{table}[t]
\centering
\caption{Wall-clock training time (hours) for each RD benchmark and operator learner, trained using fixed settings: \(T=1\), \(B=32\), and \(\Delta t=0.01\).}\label{tab:training_time_hours}
\begin{tabular}{c c c c}
\toprule
\textbf{PDE} & \textbf{NLOL (h)} & \textbf{DDOL (h)} & \textbf{DDOL-ART (h)} \\
\midrule
FN & 2.22 & 1.38 & 0.63 \\
GS & 2.28 & 1.28 & 0.71 \\
LO & 2.43 & 1.27 & 0.67 \\
\bottomrule
\end{tabular}
\end{table}

Before moving to full OOD generalization grids, we isolate the effect of the adaptive recurrent mechanism in DDOL-ART through a controlled training-dynamics study on the GS system. To ensure a fair comparison, we intentionally do not use the ID-optimal hyperparameters of each method. We fix a common configuration for all runs: batch size $B=32$, training horizon $T=1$, and operator step size $\Delta t=0.01$. We then train four models under identical architecture and optimizer settings: DDOL-ART with $n_{\mathrm{fail}}\in\{1,2,4\}$, and standard DDOL, which corresponds to $n_{\mathrm{fail}}=\infty$ (no early-exit resets). Each setting is repeated with three independent random seeds, and we report validation and test errors averaged over seeds, computed using the checkpoint that minimizes the batch-end validation loss.

At the end of each training batch, we evaluate an ID validation batch consisting of $K=8$ toroidal single-Gaussian ICs, and a fixed OOD test batch drawn from a pool of strong-shift initial-condition families: dot lattice, low-amplitude Turing noise, oriented stripes, spiral waves, broken fronts, annulus, patch, noisy Gaussian, and multi-Gaussian. To remove evaluation noise, the OOD batch is sampled once at initialization and then kept fixed throughout training. Both validation and test OOD errors are computed by autoregressively rolling out the learned flow map and measuring a worst-case (over time) MSE, which is deliberately aligned with our stability-oriented philosophy (AMAE-style evaluation). We log these batch-level metrics together with cumulative training-only wall-clock time to enable an efficiency comparison at fixed error levels.

\begin{figure}[t]
  \centering
  \includegraphics[width=\linewidth]{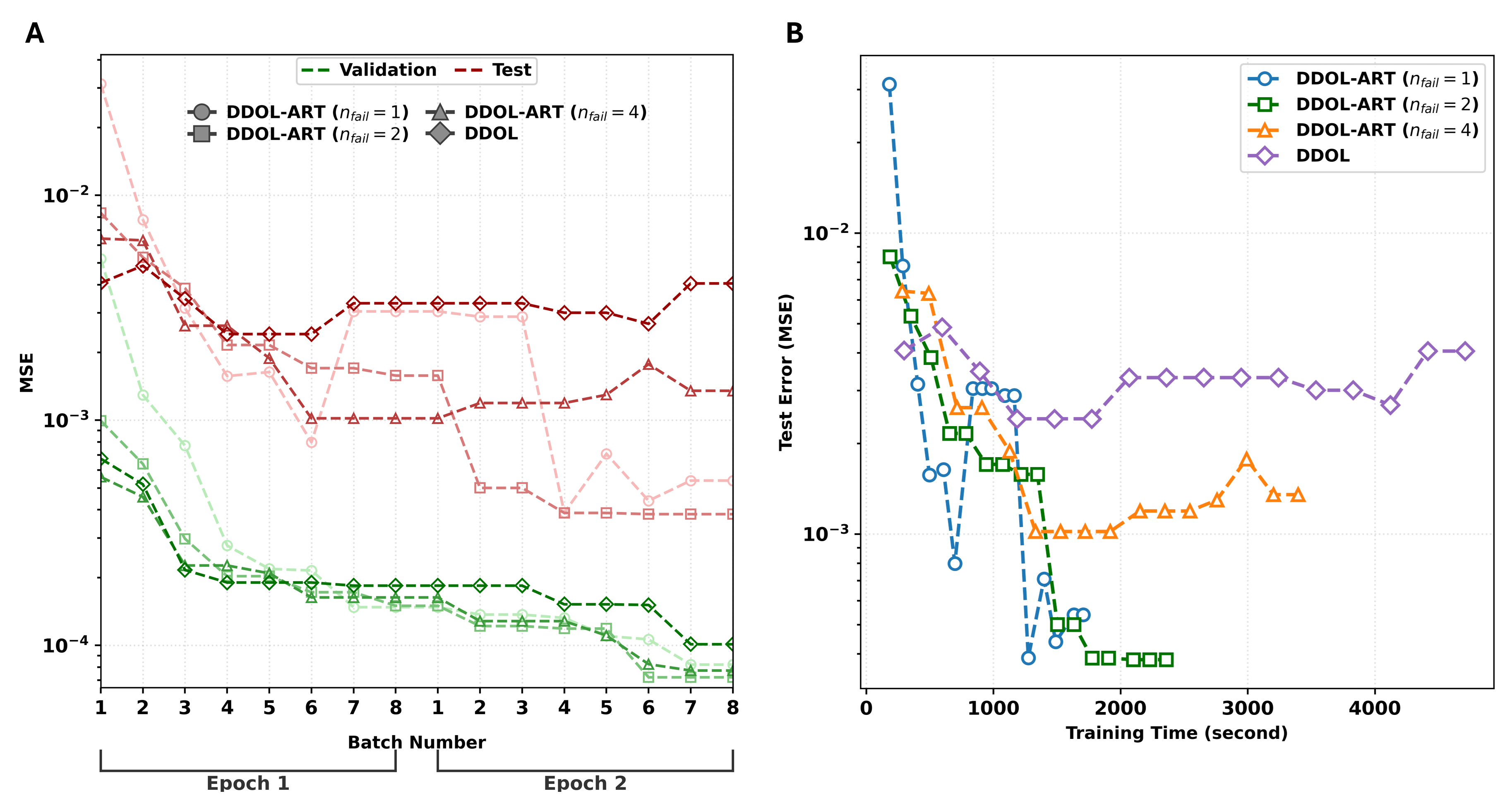}
  \caption{GS training-dynamics ablation of the DDOL-ART failure threshold $n_{\mathrm{fail}}$ under fixed hyperparameters ($B=32$, $T=1$, $\Delta t=0.01$), averaged over three random seeds. \textbf{(A)} Batch-resolved validation (ID) and OOD-test MSE during training progress including two outer epochs and eight batches for each outer epoch. Each batch includes four ID ICs during training. \textbf{(B)} Cumulative training time versus OOD-test MSE.}
  \label{fig:gs_training_dynamics}
\end{figure}

Figure~\ref{fig:gs_training_dynamics} summarizes the training traces. Figure~\ref{fig:gs_training_dynamics}A shows that all models reduce the ID validation MSE rapidly, indicating that the GS one-step map can be learned accurately under the shared hyperparameters. However, the crucial difference is how reliably this validation progress translates into OOD gains. DDOL-ART exhibits a tight coupling between the validation and OOD-test curves. When the validation error drops, the OOD-test error drops in tandem, and the late-training envelope remains bounded.
In contrast, DDOL can keep improving (or maintaining) validation performance while the OOD-test error plateaus at a substantially higher level, revealing a weak validation-test correlation link.
This gap is precisely the regime where an adaptive recurrent controller is expected to help. Rather than blindly optimizing through unproductive segments, DDOL-ART uses validation stagnation as a signal to early-exit and restart the rollout, which prevents weight updates that over-specialize to the current ID mini-batch dynamics and degrade robustness.

Figure~\ref{fig:gs_training_dynamics}B recasts the same information as OOD-test error versus cumulative training time.
DDOL-ART with $n_{\mathrm{fail}}=2$ reaches the lowest OOD-test MSE the fastest, demonstrating the best accuracy-cost trade-off.
Qualitatively, $n_{\mathrm{fail}}=1$ is often too aggressive and frequent early exits can prevent sustained improvement once the model enters a slower but productive refinement phase. Conversely, DDOL-ART with $n_{\mathrm{fail}}=4$ can be too permissive to failure in decreasing the validation error. Allowing many consecutive non-improvements increases the chance of drifting into parameter regions where additional computation no longer pays off for OOD generalization. The intermediate choice $n_{\mathrm{fail}}=2$ emerges as a robust compromise between exploiting productive optimization segments and cutting off drift that would otherwise inflate OOD error. This explains why we use $n_{\mathrm{fail}}=2$ throughout the paper.

\begin{figure}[t]
  \centering
  \includegraphics[width=\linewidth]{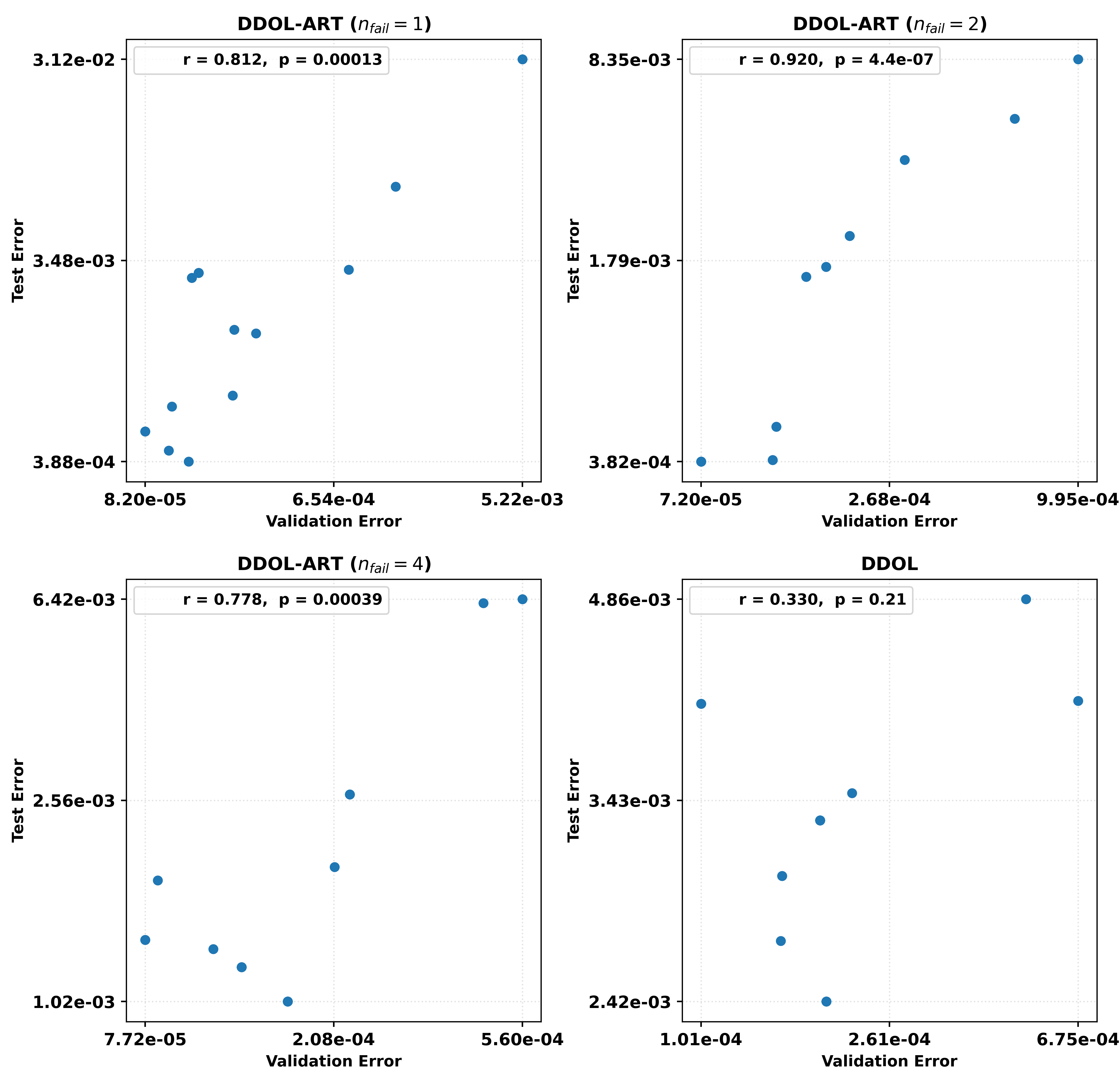}
  \caption{Batch-level validation MSE versus OOD-test MSE, together with Pearson correlation coefficients ($r$) and two-sided $p$-values.}
  \label{fig:gs_val_test_corr}
\end{figure}

Inferring the generalization ability of operator learners during the training process is a key property to avoid overfitting. Figure~\ref{fig:gs_val_test_corr} provides the Pearson correlation between batch-level validation errors and OOD-test errors evaluated after each training mini-batch. DDOL-ART yields a strong and statistically significant correlation across all tested $n_{\mathrm{fail}}$ values (with the highest correlation at $n_{\mathrm{fail}}=2$), meaning that validation improvements are predictive of OOD improvements when the adaptive controller is active. In sharp contrast, DDOL exhibits a weak and statistically insignificant correlation, implying that picking the best validation checkpoint alone is not sufficient to guarantee OOD gains when the training trajectory is not regulated. Therefore, DDOL-ART should be viewed not merely as a checkpointing heuristic for reducing the training cost, but as a training-time feedback mechanism that reshapes the optimization path to preserve robustness. These observations explain why, in the broader OOD experiments, DDOL-ART tends to maintain bounded long-horizon error envelopes on challenging GS morphologies, while conventional DDOL can be competitive yet less reliably aligned with OOD robustness.

\begin{figure}
  \centering
  \makebox[\textwidth][c]{\includegraphics[width=\textwidth]{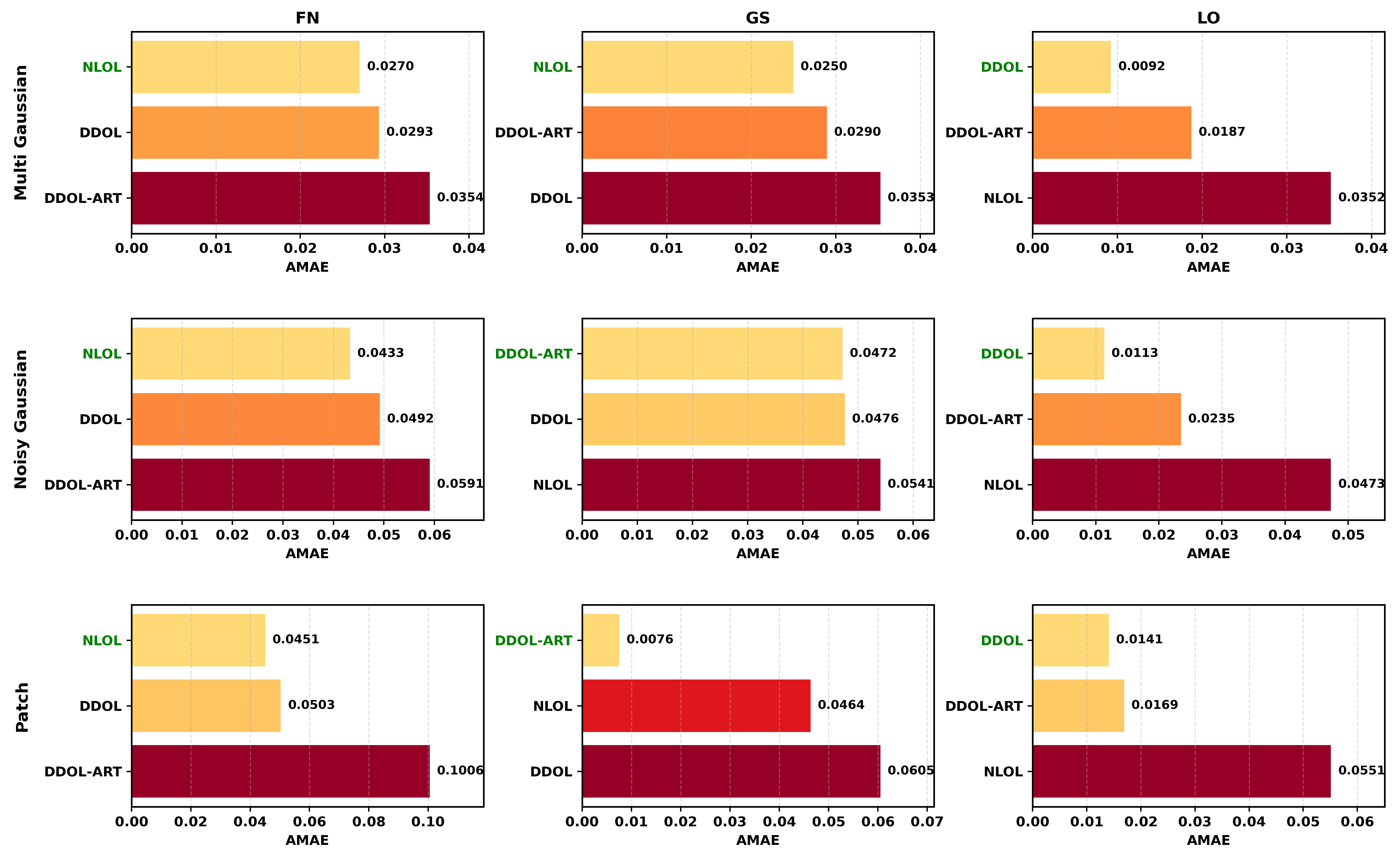}}
  \caption{Out-of-distribution performance of NLOL, DDOL, and DDOL-ART on three unseen initial-condition families: multi-Gaussian, noisy Gaussian, and patch. All models are trained only on the in-distribution single-Gaussian family using the fixed ID-selected configurations from Table~\ref{tab:model_selection}, and are evaluated by rolling out to \(T_{\text{test}}=10\) over \(10\) random seeds. Bars are sorted in ascending order, and the best method in each panel is highlighted in green.}
  \label{fig:amae_ood}
\end{figure}

A key question is whether operators trained only on the ID single-Gaussian family can generalize zero-shot to strong OOD morphology shifts without sacrificing long-horizon stability. Figure~\ref{fig:amae_ood} shows that all three approaches exhibit non-trivial zero-shot transfer to OOD initial conditions that are never observed during training, while revealing clear system-dependent strengths. In the case of the FN system, the physics-based NLOL is consistently the most accurate across the three OOD families (multi-Gaussian: $0.0270$, noisy Gaussian: $0.0433$, patch: $0.0451$), with DDOL remaining close and DDOL-ART yielding larger errors in this benchmark. Turning to the GS system, DDOL-ART provides the strongest robustness, most notably under the patch shift where it reduces AMAE to $0.0076$ compared with $0.0464$ (NLOL) and $0.0605$ (DDOL), indicating that the adaptive recurrent mechanism can dramatically suppress worst-case errors for strongly shifted morphologies. Finally, regarding the LO system, the purely data-driven DDOL attains the best AMAE across all three families (multi-Gaussian: $0.0092$, noisy Gaussian: $0.0113$, patch: $0.0141$), while DDOL-ART remains competitive and substantially improves over NLOL in each case. Even when it is not uniformly best across all systems, DDOL-ART achieves highly competitive accuracy and can deliver large gains on challenging OOD families, while requiring significantly less training time under matched settings.

\begin{figure}
  \centering
  \includegraphics[width=\textwidth]{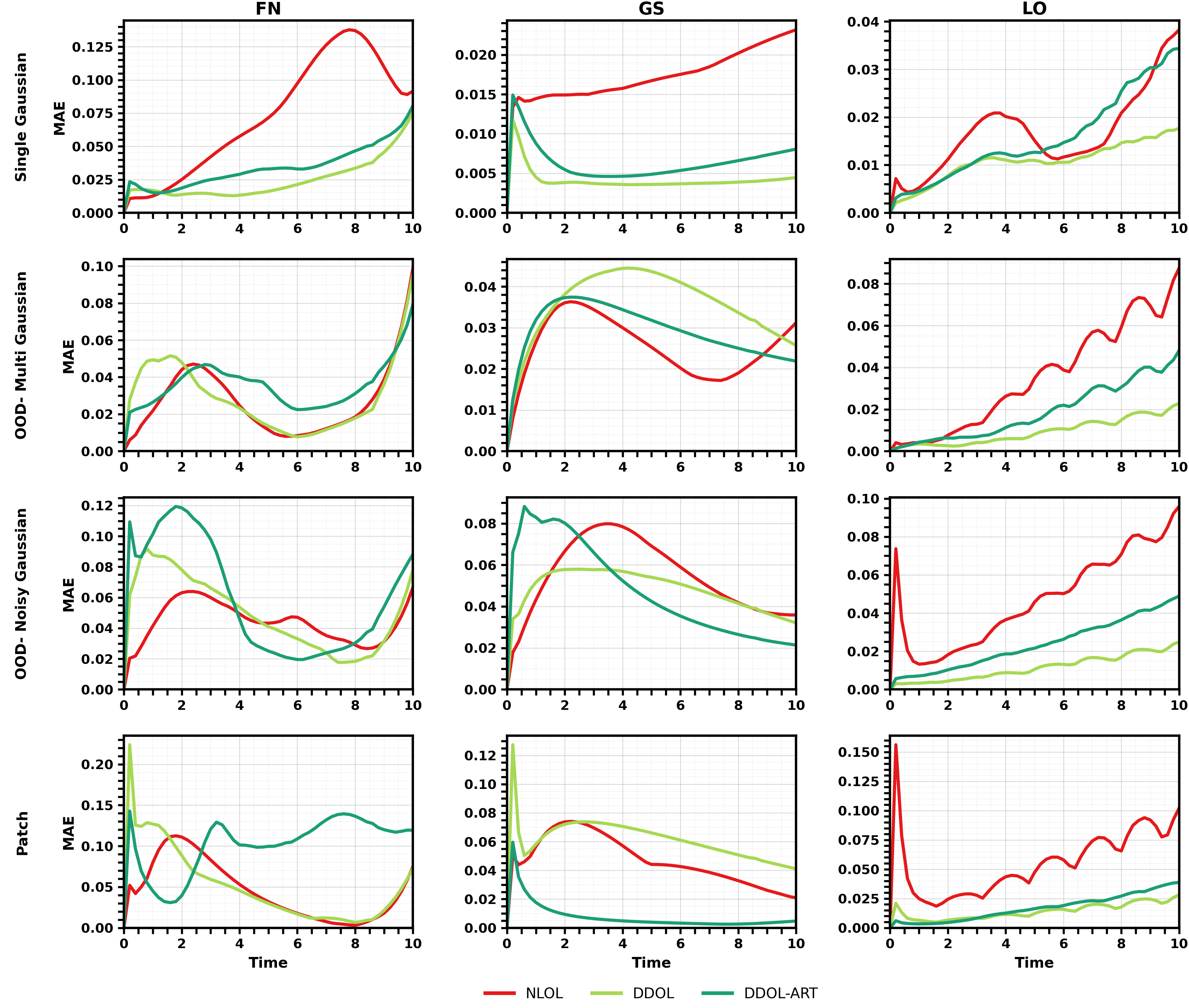}
\caption{Time-resolved MAE up to \(T_{\text{test}}=10\) for NLOL, DDOL, and DDOL-ART models. Columns show different problems FN, GS, and LO. Rows show the generalization performance of each operator learning model for ID (single Gaussian) and the OOD families: multi-Gaussian, noisy Gaussian, and patch. At each time \(t\), curves report MAE averaged over ten independently sampled initial conditions of the indicated family.}

  \label{fig:time_mae_grid}
\end{figure}

Because long-horizon stability is determined not only by the final error but also by how rapidly errors grow under autoregressive composition, it is essential to inspect time-resolved error accumulation along the rollout. Figure~\ref{fig:time_mae_grid} complements the AMAE summaries by revealing when and how errors accumulate along the rollout. On the in-distribution single-Gaussian tests, DDOL produces the lowest trajectories for GS and LO over most of \(t\in[0,10]\), while DDOL-ART remains close and typically preserves a similar late-time slope, indicating stable long-horizon propagation at the common evaluation horizon \(T_{\text{test}}=10\). NLOL, by contrast, can exhibit a more pronounced upward drift in GS and LO, which is consistent with accumulating mismatch between the learned operator and multi-step dynamics even when early-time behavior is accurate.

Across the moderate OOD families (multi-Gaussian and noisy Gaussian), the ranking becomes PDE-specific rather than uniform. In the GS case, DDOL-ART is particularly effective at controlling late-time growth, matching or improving upon DDOL and clearly outperforming NLOL in several panels, which aligns with its strong AMAE results on the most shifted GS cases. Conversely, for LO, DDOL remains the most accurate and stable across time, with DDOL-ART usually second-best and NLOL often exhibiting the largest errors. Finally, regarding FN, all three methods can stay bounded on multi-Gaussian and noisy Gaussian, but the patch family highlights a sharp separation: DDOL and NLOL decay to very small errors at late times, whereas DDOL-ART sustains a higher error envelope, explaining its larger AMAE for FN patch in Fig.~\ref{fig:amae_ood}. Overall, DDOL-ART is best viewed here as a competitive and stable alternative that is especially strong on GS-style shifts, while DDOL can be the most accurate choice on LO, and NLOL remains highly competitive on FN for several OOD families.

\begin{figure}
  \centering
  \includegraphics[width=\textwidth]{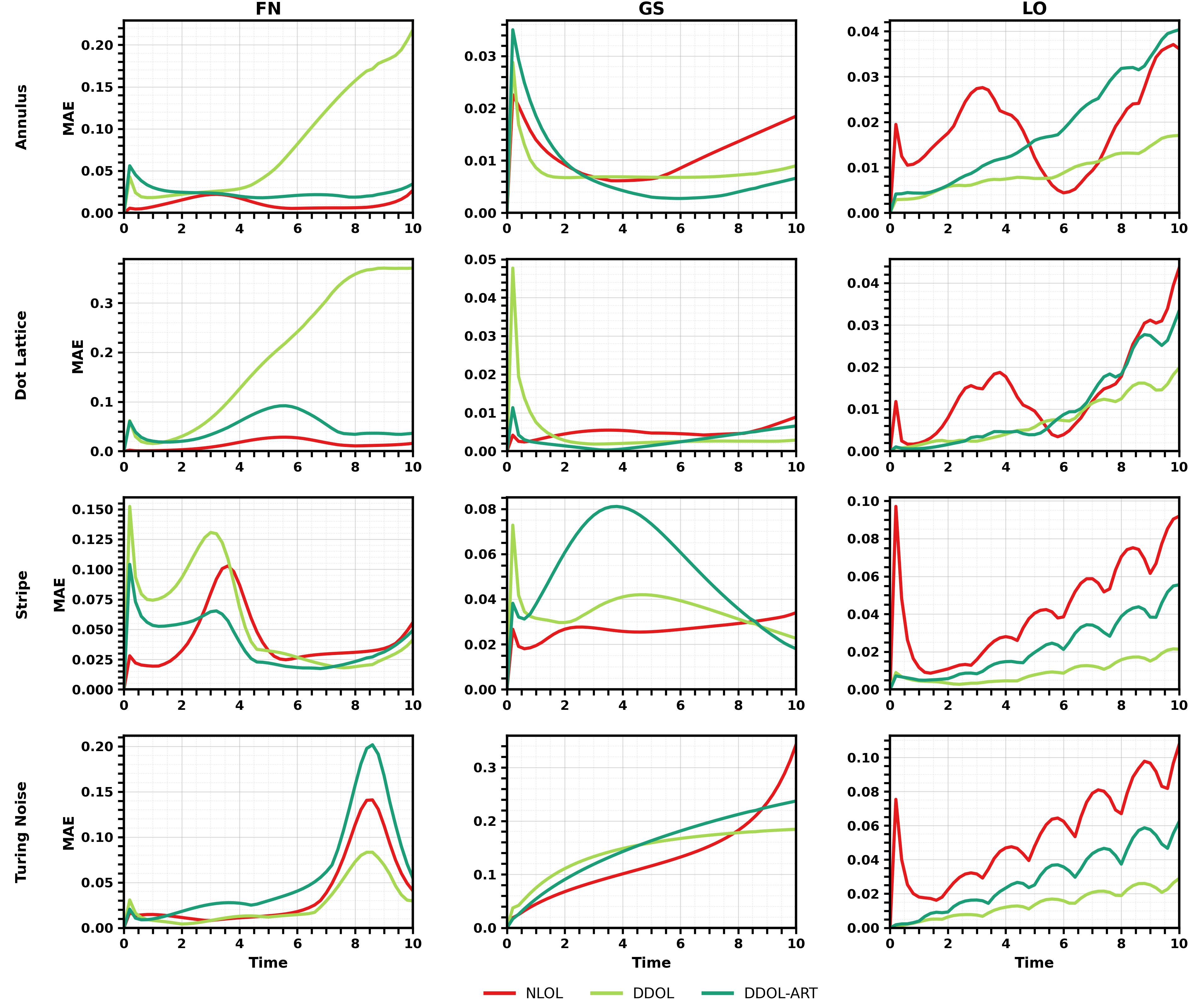}
\caption{Time-resolved MAE up to \(T_{\text{test}}=10\) for solving RD systems having strong OOD initial conditions. Columns show different problems FN, GS, and LO; rows show the OOD families: annulus, dot lattice, stripe, and Turing noise. At each time \(t\), curves report MAE averaged over ten independently sampled initial conditions of the indicated family; line colors follow the NLOL, DDOL, and DDOL-ART models.}
  \label{fig:time_mae_grid_2}
\end{figure}

Because endpoint metrics can hide transient instabilities and late-time blow-ups under strong distribution shift, we also evaluate time-resolved errors on more extreme OOD morphologies (annulus, dot lattice, stripe, Turing noise) in Figure~\ref{fig:time_mae_grid_2}. In the case of FN, plain DDOL shows clear failure modes under annulus and dot-lattice initial conditions, with monotone error growth to large values, while DDOL-ART remains bounded and substantially suppresses long-horizon drift, indicating that the adaptive recurrent schedule improves robustness to structured shifts that can destabilize free-run training. When applied to GS, all methods handle annulus and dot-lattice relatively well, but stripe and Turing noise produce larger transients.  DDOL-ART can show a higher mid-horizon peak in stripe, yet it still avoids the uncontrolled late-time growth observed for NLOL in Turing noise and stays within a bounded envelope. Finally, in the LO setting, DDOL is consistently the most stable and accurate across all four families, DDOL-ART remains competitive with modest growth, and NLOL exhibits the largest early spikes and the strongest drift in several cases, reflecting sensitivity to stiff pattern interactions.

Taken together with Table~\ref{tab:training_time_hours}, these time-resolved traces support a practical takeaway: DDOL-ART offers a strong accuracy-cost trade-off and improved robustness in several difficult shifted regimes, while DDOL can be the most accurate option when it remains stable, and NLOL can be highly effective on specific systems but is less uniformly reliable across the strongest OOD morphologies.

Having established the aggregate and time-resolved accuracy trends presented in Figs.~\ref{fig:time_mae_grid}–\ref{fig:time_mae_grid_2}, we now complement them with qualitative reconstructions under more pronounced distribution shifts. The objective is to visually evaluate whether the operators preserve the salient RD morphologies and phase relations over extended horizons. To this end, we compare the numerical reference solution against DDOL-ART on two challenging OOD families: annulus and Turing noise. Notably, the presented DDOL-ART models were trained exclusively on single Gaussian ID data with a recurrent horizon \(T=5\), yet they are evaluated zero-shot up to \(T_{\text{test}}=10\) under OOD initial conditions.

Across both families, DDOL-ART closely tracks the reference patterns from noisy or ringed ICs through the nonlinear transient and into the long-time regime. Interfaces, lobe locations, and \(u-v\) phase relations are retained, while MAE remains bounded and typically small even at \(t=10\). These visualizations are consistent with the quantitative evidence in Figs.~\ref{fig:amae_ood}–\ref{fig:time_mae_grid_2}. Although trained on a single-Gaussian initial condition family and only up to \(T=5\), DDOL-ART generalizes robustly to strong OOD ICs and longer horizons, suppressing late-time drifts that otherwise accumulate in standard data-driven rollouts. 

\begin{figure}
  \centering
  \includegraphics[width=\textwidth]{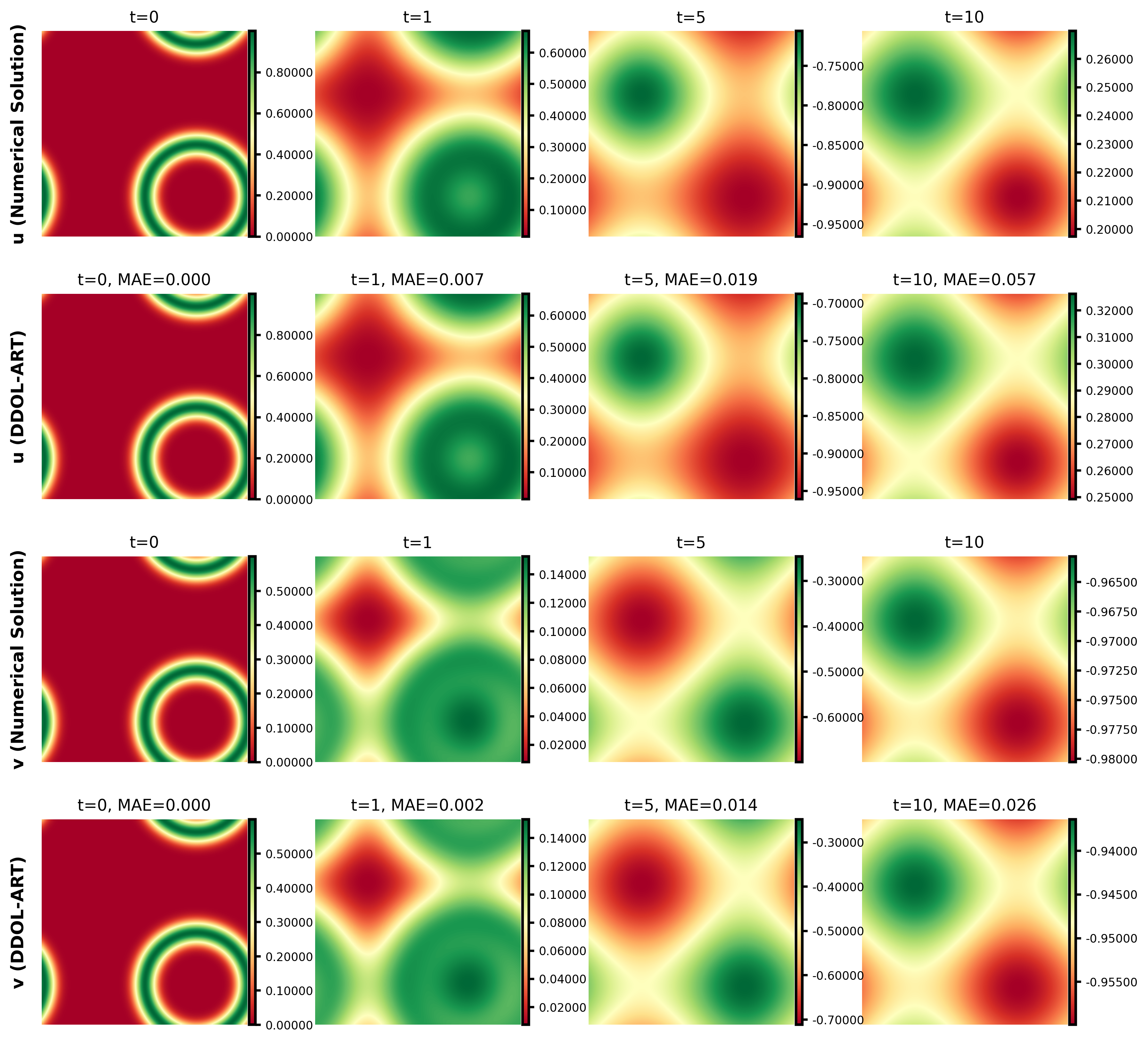}
  \caption{Qualitative OOD reconstruction on the annulus initial condition. Rows show the \(u\) and \(v\) components for the numerical solution (reference) and DDOL-ART; columns are snapshots at representative times up to \(T=10\) (\(t\in\{0,1,5,10\}\)). Each panel reports the instantaneous MAE at that time. }
  \label{fig:qual_ood_annulus}
\end{figure}

\begin{figure}
  \centering
  \includegraphics[width=\textwidth]{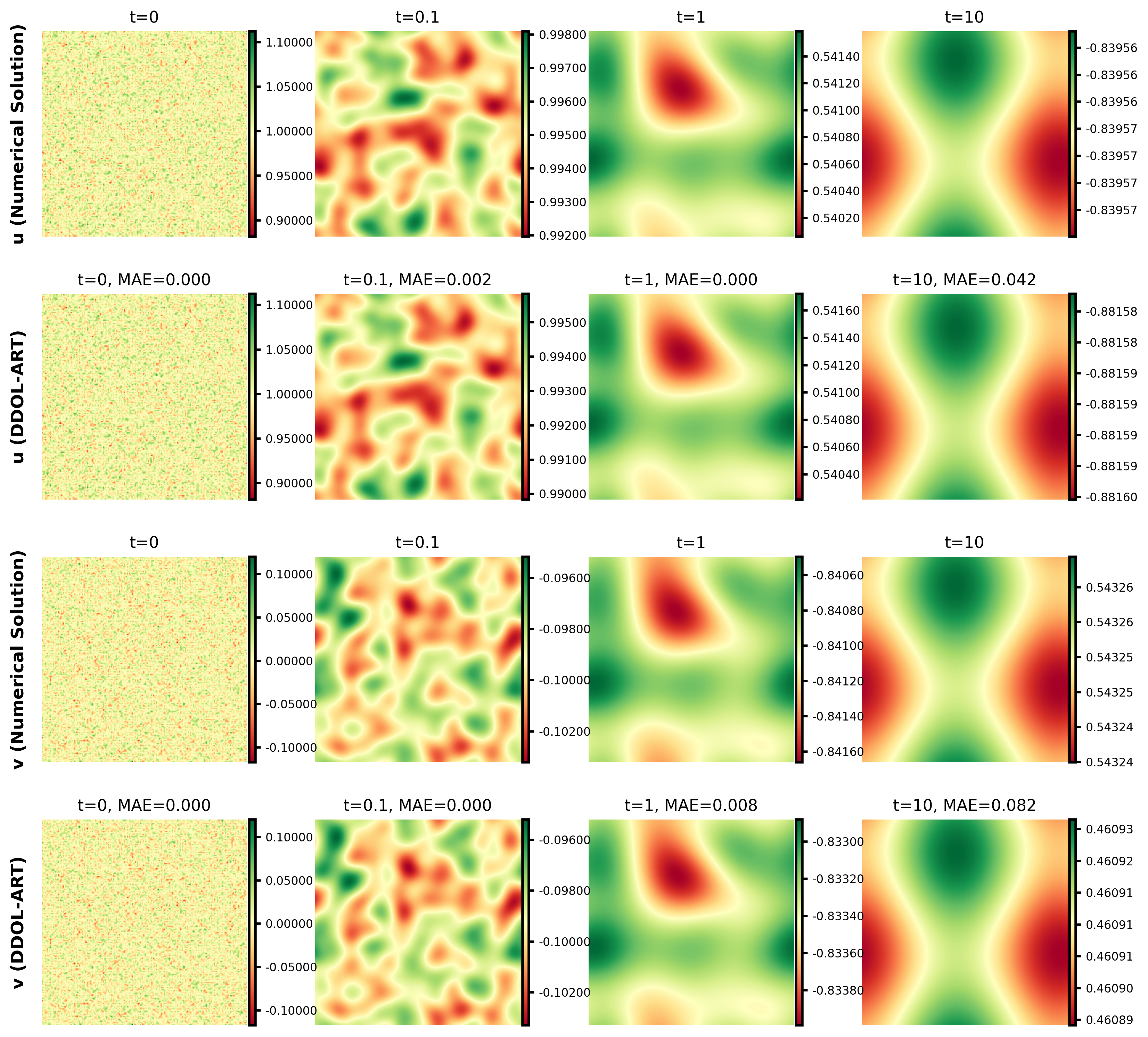}
  \caption{Qualitative OOD reconstruction on Turing noise initial condition. Rows show the \(u\) and \(v\) components for the numerical solution (reference) and DDOL-ART; columns are snapshots at representative times up to \(T=10\) (\(t\in\{0,0.1,1,10\}\)). Each panel reports the instantaneous MAE at that time.}
  \label{fig:qual_ood_turing}
\end{figure}
\

\section{Discussion and Conclusions}\label{sec:discussion}

Learning accurate long-horizon surrogates for time-dependent PDEs is limited by the mismatch between training and deployment. When one-step predictors are trained near the data manifold, for example by teacher forcing or local residual minimization, they may appear accurate at short times but drift under autoregressive composition. In that setting, each step conditions on the model’s own slightly off-manifold states. This exposure-bias mechanism is particularly consequential in reaction-diffusion dynamics, where nonlinear reactions and diffusion can turn small phase and amplitude errors into visible morphological drift over long rollouts. Recent work has therefore emphasized recurrent, inference-aligned supervision as a principled route to stability, with analyses showing improved worst-case error growth compared with purely one-step objectives \citep{geneva2020modeling,Ye2025_RNO}.

A common expectation is that physics-based training objectives, especially weak-form or fully discrete residual losses, enhance generalization and reliability in operator learning, sometimes even with reduced dependence on labeled trajectories \citep{Li2021_PINO,Kovachki2023,Geng2024}. Our study refines this narrative by isolating the role of the training strategy that governs how the learned flow map is composed during optimization. We compared NLOL, which minimizes a fully discrete CN--FD2 residual without trajectory labels, DDOL, which uses supervised SSP--RK3+FD2 targets under free-run recurrent supervision, and DDOL-ART, which augments DDOL with an adaptive recurrent controller that monitors validation progress at milestones and early-exits unproductive rollout segments. All methods are assessed under the same stability-oriented evaluation. Every trained model is rolled out to a fixed horizon, and performance is measured by worst-case error summaries and time-resolved MAE traces.

Our experiments show that stable long-horizon behavior is learned most reliably when the flow map is trained under longer free-run compositions and when the optimization trajectory is actively regulated. In-distribution selection consistently favors recurrent training that exposes the model to multi-step compounding during optimization, which shapes a one-step operator that remains accurate under repeated composition rather than only near the data manifold. Building on the robust recurrent strategy provided by \citep{Geng2024}, we introduce an adaptive recurrent controller and find that DDOL-ART shortens training substantially relative to residual-based training while remaining competitive on both in-distribution and shifted tests. The key mechanism is a tighter coupling between validation progress and robustness under distribution shift. The controller uses validation stagnation as feedback to stop unproductive rollout segments and redirect updates, which improves the efficiency of recurrent optimization and reduces drift toward parameter regions that degrade generalization. Physics-based residual objectives can still be highly effective in specific regimes, but our results refine the common narrative by showing that a carefully designed, inference-aligned recurrent strategy with lightweight adaptivity can be a primary driver of long-horizon stability and shift-robust generalization for flow-map operator learning.

The GS training-dynamics ablation clarifies why adaptivity helps. Under fixed hyperparameters, DDOL-ART exhibits a tight coupling between in-distribution validation progress and OOD-test improvements (Fig.~\ref{fig:gs_training_dynamics}A), and the validation to OOD correlation becomes strong and statistically significant across failure thresholds (Fig.~\ref{fig:gs_val_test_corr}). In contrast, standard DDOL shows a weak, non-predictive linkage. This indicates that DDOL-ART acts as a training-time feedback controller. It uses validation stagnation as a signal to redirect optimization, avoids spending compute in segments that do not translate to robustness, and improves the accuracy and time trade-off. Mechanistically, this aligns with the core stability objective of recurrent training. The model is repeatedly exposed to its own predicted states and is incentivized to learn corrective dynamics that suppress compounding error under composition. The adaptive early-exit policy then regularizes the optimization path by preventing prolonged drift that can arise when free-run training over-specializes to particular mini-batch dynamics or enters parameter regions that degrade OOD behavior.

Taken together, our results support the view that robust recurrent supervision, augmented by lightweight adaptivity, can be as decisive for long-horizon stability and distribution-shift robustness as the choice of physics penalty itself. Physics-based residual objectives remain valuable and can be excellent in certain regimes, as seen in FN, but they do not by themselves guarantee well-calibrated multi-step compositions across diverse OOD morphologies. Conversely, a carefully designed data-driven recurrent strategy can deliver stable operators that generalize zero-shot from a single in-distribution initial-condition family to substantial morphology shifts, while also reducing training time. Because DDOL-ART is formulated as a data-driven recurrent training protocol rather than a PDE-specific construction, it also suggests a broader implication. Stable flow-map learning with feedback-controlled recurrent supervision may transfer naturally to dynamical-systems forecasting from data in settings where explicit governing equations are unavailable, such as generic spatiotemporal time-series systems.

Our study is limited to coupled two-field reaction-diffusion systems on periodic domains, and it assumes dense, fully observed state trajectories generated by a known high-fidelity solver. These choices isolate the effect of training strategy, but they leave open how DDOL-ART behaves under sparse observations, partial-state measurements, heterogeneous domains and boundary conditions, and broader PDE families with different stiffness and multiscale structure. Future work will extend the evaluation to additional evolution PDE classes and to non-periodic geometries. We will also relax the dense-data assumption by integrating DDOL-ART with sparse-sensor or masked supervision and with observation operators. 

\section*{Declarations}
\noindent
\textbf{Competing interests} The authors declare that they have no known competing interests.\\
\textbf{Code, data and materials availability} Code, materials and data can be provided if requested.\\

\newpage
\appendix

\section*{Appendix A. Initial Condition Distributions}

All initial conditions are posed on the periodic torus $\Omega=\mathbb{T}^2=[0,1]^2$ with wrap-around (minimum-image) distance
\[
d_{\mathbb{T}^2}^2(\mathbf{x},\mathbf{c})
=\min\!\big(|x-c_x|,\,1-|x-c_x|\big)^2
+\min\!\big(|y-c_y|,\,1-|y-c_y|\big)^2 .
\]

\textbf{Training distribution:} A single toroidal Gaussian with random center and width:
\[
u_0(\mathbf{x})=\exp\!\Big(-\frac{d_{\mathbb{T}^2}(\mathbf{x},\mathbf{c})^2}{2\sigma^2}\Big),
\qquad \mathbf{c}\sim\mathcal{U}([0,1]^2),\;\; \sigma\in[0.05,\,0.20].
\]
For two-component systems $(u,v)$, independent draws are used unless noted.

\textbf{OOD: Multi-Gaussian:} A finite superposition of Gaussians with random centers and widths:
\[
n\sim\mathcal{U}_{\mathbb{Z}}\{2,\ldots,5\},\quad 
\mathbf{c}_k\sim\mathcal{U}([0,1]^2),\quad 
\sigma_k\in[\sigma_{\min},\sigma_{\max}],
\]
\[
u_0(\mathbf{x})=\frac{1}{n}\sum_{k=1}^{n}
\exp\!\Big(-\frac{d_{\mathbb{T}^2}(\mathbf{x},\mathbf{c}_k)^2}{2\sigma_k^2}\Big).
\]

\textbf{Noisy Gaussian:} A single toroidal Gaussian perturbed by i.i.d. spatial white noise:
\[
u_0(\mathbf{x})=\exp\!\Big(-\frac{d_{\mathbb{T}^2}(\mathbf{x},\mathbf{c})^2}{2\sigma^2}\Big)+\eta(\mathbf{x}),
\quad \mathbf{c}\sim\mathcal{U}([0,1]^2),\ \sigma\in[\sigma_{\min},\sigma_{\max}],
\]
\[
\eta(\mathbf{x})\stackrel{\text{i.i.d.}}{\sim}\mathcal{N}(0,\sigma_\eta^2),\qquad 0<\sigma_\eta\ll1.
\]
For two-component systems, independent noise realizations are applied to each component.

\textbf{Patch:} An axis-aligned periodic rectangle of ones on a zero background:
\[
\mathbf{c}\sim\mathcal{U}([0,1]^2),\quad
w\sim\mathcal{U}([w_{\min},w_{\max}]),\quad
h\sim\mathcal{U}([h_{\min},h_{\max}]),
\]
set $u_0{=}1$ on the $w{\times}h$ rectangle centered at $\mathbf{c}$ (with toroidal wrap) and $u_0{=}0$ elsewhere.

\textbf{Turing noise:} Let $(u^*,v^*)$ be a spatially homogeneous equilibrium. We define spatial white noise on the grid as an i.i.d. Gaussian field with zero mean and variance $\sigma_\eta^2$ at each grid point, independent across components:
\[
\eta_u(\mathbf{x}),\,\eta_v(\mathbf{x})\stackrel{\text{i.i.d.}}{\sim}\mathcal{N}(0,\sigma_\eta^2),\qquad
u_0(\mathbf{x})=u^*+\eta_u(\mathbf{x}),\ \ v_0(\mathbf{x})=v^*+\eta_v(\mathbf{x}),
\]
with $0<\sigma_\eta\ll1$. This perturbs all wavelengths at $t{=}0$; subsequent linear growth selects a finite band.

\textbf{Annulus:} Choose a random center $\mathbf{c}\sim\mathcal{U}([0,1]^2)$, a target radius $r_0\sim\mathcal{U}([r_{0,\min},r_{0,\max}])$, and thickness $s\sim\mathcal{U}([s_{\min},s_{\max}])$. With toroidal distances $d_x,d_y$ from $\mathbf{c}$ and $r(\mathbf{x})=\sqrt{d_x^2+d_y^2}$, set
\[
u_0(\mathbf{x})=\exp\!\Big(-\tfrac{(r(\mathbf{x})-r_0)^2}{2s^2}\Big).
\]

\textbf{Oriented stripes:} Draw orientation $\phi\sim\mathcal{U}([0,2\pi])$ and integer spatial frequency $f\sim\mathcal{U}_{\mathbb{Z}}\{f_{\min},\ldots,f_{\max}\}$. With $s(\mathbf{x})=\cos\phi\,x+\sin\phi\,y$,
\[
u_0(\mathbf{x})=\tfrac{1}{2}\Big(1+\sin\big(2\pi f\,s(\mathbf{x})\big)\Big).
\]

\textbf{Dot lattice:} Choose lattice counts 
$n_x,n_y\sim\mathcal{U}_{\mathbb{Z}}\{4,\ldots,7\}$ and set cell widths 
$s_x=1/n_x$, $s_y=1/n_y$. Let the Gaussian width be 
$\sigma=0.12\,\min(s_x,s_y)$. For each cell $(i,j)$, define a jittered toroidal center using the fractional part operator $\operatorname{frac}(\cdot)$:
\[
\mathbf{c}_{ij}=\Big(
\operatorname{frac}\!\big((i+\tfrac{1}{2}+\delta^{(x)}_{ij})\,s_x\big),\;
\operatorname{frac}\!\big((j+\tfrac{1}{2}+\delta^{(y)}_{ij})\,s_y\big)
\Big),
\quad \delta^{(x)}_{ij},\delta^{(y)}_{ij}\sim\mathcal{U}([0,0.2]).
\]
and set the (unnormalized) field as a sum of toroidal Gaussians
\[
\tilde{g}(\mathbf{x})=\sum_{i=0}^{n_x-1}\sum_{j=0}^{n_y-1}
\exp\!\Big(-\frac{d_{\mathbb{T}^2}(\mathbf{x},\mathbf{c}_{ij})^2}{2\sigma^2}\Big),
\qquad 
g(\mathbf{x})=\frac{\tilde{g}(\mathbf{x})}{\max_{\mathbf{z}\in\Omega}\tilde{g}(\mathbf{z})}.
\]
We take $u_0(\mathbf{x})=g(\mathbf{x})$. 
For two-component systems aimed at spot regimes, we initialize with a small anti-phase contrast
\[
(u_0(\mathbf{x}),\,v_0(\mathbf{x}))=\big(g(\mathbf{x}),\ \gamma\,(1-g(\mathbf{x}))\big),
\]
with \(\gamma=0.1\) and we clamp each component to $[0,1]$.







\bibliography{sn-bibliography}

\end{document}